\documentclass[letterpaper,10pt,journal,twoside]{IEEEtran}
\usepackage{amsmath,amsfonts}
\usepackage{algorithmic}
\usepackage{algorithm}
\usepackage{array}
\usepackage[caption=false,font=normalsize,labelfont=sf,textfont=sf]{subfig}
\usepackage{textcomp}
\usepackage{stfloats}
\usepackage{url}
\usepackage{verbatim}
\usepackage{graphicx}
\usepackage{cite}
\usepackage{booktabs} % Add it by me 
\usepackage{xcolor}
 \newcommand{\change}[1]{{#1}}
\usepackage{tikz}
\newcommand\copyrightnotice{%
  \begin{tikzpicture}[remember picture,overlay]
    \node[anchor=south,yshift=10pt] at (current page.south) {%
      \parbox{\dimexpr\textwidth-\fboxsep}{%
        \tiny \centering
        Accepted for publication in IEEE Robotics and Automation Letters (RA-L).\\
        \textcopyright~2026 IEEE. Personal use of this material is permitted. Permission from IEEE must be obtained for all other uses, in any current or future media, including reprinting/republishing this material for advertising or promotional purposes, creating new collective works, for resale or redistribution to servers or lists, or reuse of any copyrighted component of this work in other works.\\
        DOI: 10.1109/LRA.2026.11675828
      }%
    };
  \end{tikzpicture}%
}

\begin{document}

\title{Odometer-Agnostic Drift Correction\\ Using OpenStreetMap Lane Geometry}

% \thanks{Manuscript received April 19, 2026; revised June 12, 2026.}}

\author{Joaquin~Caballero$^{1}$,
        Emilio~Garcia-Fidalgo$^{2}$,
         Alberto~Ortiz$^{2}$,~\IEEEmembership{Member,~IEEE,},
        and~Jarno~Ralli$^{3}$%

\thanks{Manuscript received: April 13, 2026; Revised: July 14, 2026; Accepted: August 11, 2026.
This paper was recommended for publication by Editor Javier Civera upon evaluation of the Associate Editor and Reviewers comments.}%

\thanks{$^{1}$J. Caballero is with Distance Technologies Oy, Helsinki, Finland
({\tt\small joaquinecc@gmail.com}).}%
\thanks{$^{2}$E. Garcia-Fidalgo and A. Ortiz are with the Department of
Mathematics and Computer Science, Institute of Artificial Intelligence,
University of the Balearic Islands, 07122 Palma, Spain
({\tt\small emilio.garcia@uib.es}, {\tt\small alberto.ortiz@uib.es}).}%
\thanks{$^{3}$J. Ralli was with Distance Technologies Oy, Helsinki, Finland
({\tt\small jarno@ralli.fi}).\\
Digital Object Identifier (DOI): see top of this page.}%
}

    % \markboth{IEEE ROBOTICS AND AUTOMATION LETTERS,~Vol.~11, No.~6, June~2026}%
\markboth{IEEE ROBOTICS AND AUTOMATION LETTERS. ACCEPTED AUGUST, 2026}%
{Caballero Caballero \MakeLowercase{\textit{et al.}}: Odometer-Agnostic Drift Correction Using OpenStreetMap Lane Geometry}

% \IEEEpubid{0000--0000~\copyright~2026 IEEE}

\maketitle
\copyrightnotice

\begin{abstract}
Despite significant progress in odometry estimation, long-term drift remains a fundamental limitation of incremental pose integration, especially in large-scale or loop-free environments. Existing map-assisted methods can reduce drift, but often depend on dense maps, sensor-specific processing, or complex matching pipelines. We propose a lightweight open-source, odometry-agnostic correction method that aligns short trajectory segments to OpenStreetMap (OSM) lane centerlines. By formulating drift correction as a direct alignment between recent odometry and sparse lane geometry, the method enables efficient online operation without dense priors or expensive preprocessing. Experiments with LiDAR and visual odometry backends demonstrate consistent improvements, with particularly strong gains under severe drift.
\end{abstract}


\section{Introduction}
    \IEEEPARstart{O}{dometry} is a fundamental component of mobile robotics and navigation systems. It provides continuous pose estimates from onboard sensors such as LiDARs, cameras, and inertial measurement units (IMUs). However, even state-of-the-art LiDAR and Visual Odometry (VO) systems accumulate drift over long trajectories~\cite{LOAMZhang2014RSS,garcia2022liodom,orbslam3,basalt,kissicp}. This drift becomes particularly problematic in large-scale deployments where loop closures are sparse and global positioning signals may be unreliable.

    To mitigate drift, map-assisted localization techniques incorporate prior map information as global geometric references. In particular, High-Definition (HD) maps have been widely adopted for autonomous-driving localization because they are designed to provide high-precision geometric and semantic information, including lane geometry and 3D environmental structure~\cite{hdmap_survey2023}. Several works demonstrate the effectiveness of HD-map-based localization in improving accuracy and global consistency~\cite{ebrahimi2022high,hdmap_survey2023,hdmap_review2026,hdmap_semantic2021}. However, HD maps are expensive to acquire, difficult to maintain at large scale, and typically require extensive preprocessing pipelines and dedicated infrastructure.

    In contrast, OpenStreetMap (OSM) provides a freely available and globally scalable mapping resource~\cite{OpenStreetMap}. Although less precise than HD maps, OSM still captures meaningful geometric structure, including road topology and lane-level information. This naturally leads to the question of whether lightweight, open-source lane geometry provides sufficient constraints to mitigate odometry drift without the complexity and cost of HD map infrastructures.

    Motivated by these limitations, we propose a lightweight, odometry-agnostic drift correction framework that relies solely on sparse lane centerlines from OpenStreetMap (OSM). Instead of dense maps or global graph matching, we formulate drift correction as a direct alignment between short trajectory segments and lane geometry. This enables efficient real-time operation with minimal preprocessing while remaining robust under substantial accumulated drift.
  
    Unlike approaches that require dense priors such as 3D point clouds or building meshes~\cite{rl2a-ballardini2021vehicle,rl5a-kulmer2025openlidarmap}, or methods that rely  on complex road graph construction and large-scale probabilistic matching~\cite{rl4b-cheng2021graph}, our approach demonstrates that simple geometric constraints derived from lane centerlines are sufficient to significantly reduce drift in large-scale scenarios. The framework avoids brute-force map matching, full 3D registration, and large-scale graph optimization, while preserving accurate trajectories over long routes.

    The main contribution of this paper is a lightweight, open-source map-based drift correction \change{pipeline}\footnote{\change{The source code and implementation are publicly available at:\\
\url{https://github.com/Joaquinecc/icp_trajectory_alignment_osm}}}
that leverages freely available OSM lane data, eliminating the need for costly HD maps.The method is odometer-agnostic and can refine trajectories from any pose estimation source (LiDAR, visual odometry, wheel-IMU, etc.). It scales to large geographic areas through a compact representation based on lane centerline points and requires minimal preprocessing, relying solely on raw lane data without road graph construction or dense map generation. Despite its simplicity, the approach remains robust under substantial accumulated drift in large-scale environments.

    To demonstrate generality, we evaluate the proposed correction on multiple odometry backends with different sensing modalities and drift characteristics, including LiDAR and visual odometry. Apart from evaluating our method on KITTI Odometry and KITTI-360, we compare against recent OSM-assisted odometry drift-correction methods that use KISS-ICP as their baseline, as well as against TOM-Odometry, which is the closest road-geometry-based approach reporting ORB-SLAM3 results. This allows us to evaluate not only whether the proposed method improves different odometry sources, but also whether a lightweight lane-centerline alignment strategy can remain competitive with more specialized OSM-based localization and drift-correction pipelines.

\section{Related Work}
    \label{sec:related_work}
    
    Map-assisted odometry and localization methods differ mainly in the type of prior map information they use to constrain drift and improve global consistency. Existing approaches can be broadly grouped into two categories: \emph{shape or descriptor matching-based} and \emph{lane or road-geometry  matching-based}. Shape-based methods align sensor observations with map structures such as buildings, boundaries, or semantic descriptors, and are commonly associated with LiDAR-based localization. Lane-based methods instead constrain the estimated trajectory using road or lane geometry extracted from maps such as OpenStreetMap (OSM), which is the category most closely related to our work.
    
    \subsection{Shape and Descriptor Matching Approaches}
    
    Several works exploit OSM building footprints or road/building structures as global references for LiDAR localization. Yan et al.~\cite{8870918} propose a global localization method that converts OSM road and building information into compact 4-bit semantic descriptors encoding road intersections and building gaps, which are then matched with descriptors extracted from 3D LiDAR scans within a Monte Carlo Localization (MCL) framework. Ballardini et al.~\cite{rl2a-ballardini2021vehicle} use stereo images and a convolutional network to obtain building-segmented point clouds, which are matched against 3D building models generated by extruding OSM footprints. Similarly, Cho et al.~\cite{9716862} precompute descriptors from OSM building geometry at road-center locations and match them against descriptors extracted from semantically segmented LiDAR building points, enabling global localization without a prior LiDAR map or base odometry.
    
    Other methods formulate the problem as LiDAR-to-OSM alignment. OSM-SLAM~\cite{rl3a-frosi2023osm} extends ART-SLAM~\cite{frosi2022art} by building a 2D representation of buildings from LiDAR scans and aligning it with OSM building outlines using iterative closest point (ICP), providing global constraints for pose graph optimization. Kurda et al.~\cite{rl1a-kurda2024reducing} reduce LiDAR odometry drift by projecting point clouds onto the $XY$-plane, extracting building-related structures, and aligning the local map with OSM building data. Li et al.~\cite{10508124} perform LiDAR--OSM global initialization by generating virtual 2D LiDAR-like descriptors from OSM building boundaries at candidate road nodes and matching them to building features extracted from the current LiDAR scan. More recently, Li et al.~\cite{11367659} extend this idea with odometry assistance: OSM-based building-boundary descriptors are matched to LiDAR descriptors to obtain candidate global positions, which are then associated over time with the local odometry trajectory to estimate an $\mathrm{SE}(2)$ alignment. OpenLiDARMap~\cite{rl5a-kulmer2025openlidarmap} further integrates OSM building footprints and surface models into a pose-graph framework, combining scan-to-map and scan-to-scan constraints to generate globally consistent point-cloud maps. \change{Lee and Ryu~\cite{10506654}  demonstrate high-accuracy localization from public 2-D maps through a multi-stage pipeline combining LiDAR--IMU mapping, Siamese/NetVLAD place recognition, particle filtering, template matching, and voxelized GICP. This additional complexity comes at a significant computational cost, with their implementation operating at about $7$~Hz, below the $10$~Hz LiDAR rates.}
    
    Although these methods can provide accurate global alignment and substantial drift reduction, they are often dependent on the presence and quality of buildings or other urban structures. Consequently, their applicability may be reduced in rural, suburban, or open environments with limited façades. In addition, many of them require LiDAR sensing, semantic segmentation, descriptor generation, simulated scans, or dense point-cloud registration, which increases preprocessing and computational complexity. In contrast, our method does not rely on building geometry, dense priors, or LiDAR-specific processing, and instead uses only lightweight lane centerlines extracted from OSM.

    \subsection{Lane and Road-Geometry Matching Approaches}
    
    Lane and road-network matching methods are more closely related to our approach because they use the road layout itself as a geometric prior. Early works focused on aligning visual odometry or proprioceptive trajectories with OSM road maps~\cite{rl1b-floros2013openstreetslam,7152950,rl2b-wang2014snap}. OpenStreetSLAM~\cite{rl1b-floros2013openstreetslam} uses an MCL framework with Chamfer distance to compare VO trajectory segments against OSM road geometry and select the best alignment hypothesis. Brubaker et al.~\cite{7152950} constrain VO-based vehicle poses to an OSM road graph, reducing drift by enforcing consistency with the road network. O-SNAP~\cite{rl2b-wang2014snap} divides an odometry trajectory into straight segments and snaps them onto a 2D road map through an optimization-based route-identification process. While effective, these methods often require global or brute-force matching over candidate road paths, which can become expensive as the map size increases.
    
    Other approaches combine road geometry with additional sensing or graph-based inference. Ruchti et al.~\cite{rl3b-ruchti2015localization} classify 3D LiDAR observations into road and non-road regions and probabilistically align the resulting classification grid with OSM road geometry using a dedicated sensor model within MCL. Suger and Burgard~\cite{7989169} use OSM road and trail geometry as navigation waypoints and correct their local placement by scoring candidate poses against a LiDAR-based semantic road grid using Markov Chain Monte Carlo (MCMC). Cheng et al.~\cite{rl4b-cheng2021graph} convert OSM-derived road maps into a Heading-Length Graph, probabilistically match heading-length sequences extracted from the vehicle trajectory for global localization, and use the resulting alignment to bound odometric drift. Xu et al.~\cite{rl5b-xu2023road} introduce point-line constraints from road-network maps into a pose-graph formulation to suppress VO drift, but they assume that the estimated trajectory remains close to the correct road segment and cannot recover globally from large errors.

    TOM-Odometry~\cite{9940585} is closely related to our work, using OSM road geometry to reduce odometry drift. It initializes the vehicle pose through road-shape matching and applies online corrections using closest-point and B-spline-based associations fused with odometry via a Kalman filter. However, it requires sufficiently low initial drift, and its local nearest-neighbor-style road associations may become unreliable in dense networks or when large drift brings the trajectory closer to an incorrect road segment.
    
    More recent map-based localization methods also exploit road constraints probabilistically. Przewodowski et al.~\cite{Przewodowski_SantosOsorio_GrassiJunior_2024} use OSM driveable road areas and elevation offsets to weight particles in an MCL framework, indirectly reducing drift by penalizing poses outside valid road regions. However, this does not directly align the trajectory to lane geometry and can remain ambiguous on wide or parallel roads, in regions with incomplete OSM data, or under large accumulated drift.
    
    In contrast to these methods, our approach performs continuous online correction by directly aligning recent odometry segments to OSM-derived lane centerlines. It does not require dense LiDAR maps, building models, semantic segmentation, simulated sensor scans, global road-graph search, or a dedicated initialization stage. Instead, it uses a compact KD-tree representation of lane centerline points, direction-aware correspondence filtering, and trimmed $\mathrm{SE}(2)$ alignment in a sliding window. This makes the method lightweight, odometry-agnostic, and applicable to both LiDAR and visual odometry backends while remaining scalable to large map regions.

\section{Method}
    \label{sec:method}

    Our approach can be understood as a continuous 2D alignment procedure that corrects odometry drift by registering variable-length segments of the estimated trajectory to lane-level map points extracted from OSM. Specifically, we use lane centerlines as lightweight geometric landmarks and solve for an incremental rigid transform in $\text{SE}(2)$ that refines the odometry output over time.

\begin{figure}[!b]
    \centering

    \includegraphics[width=0.65\columnwidth,clip=true,trim=0 0 0 100]{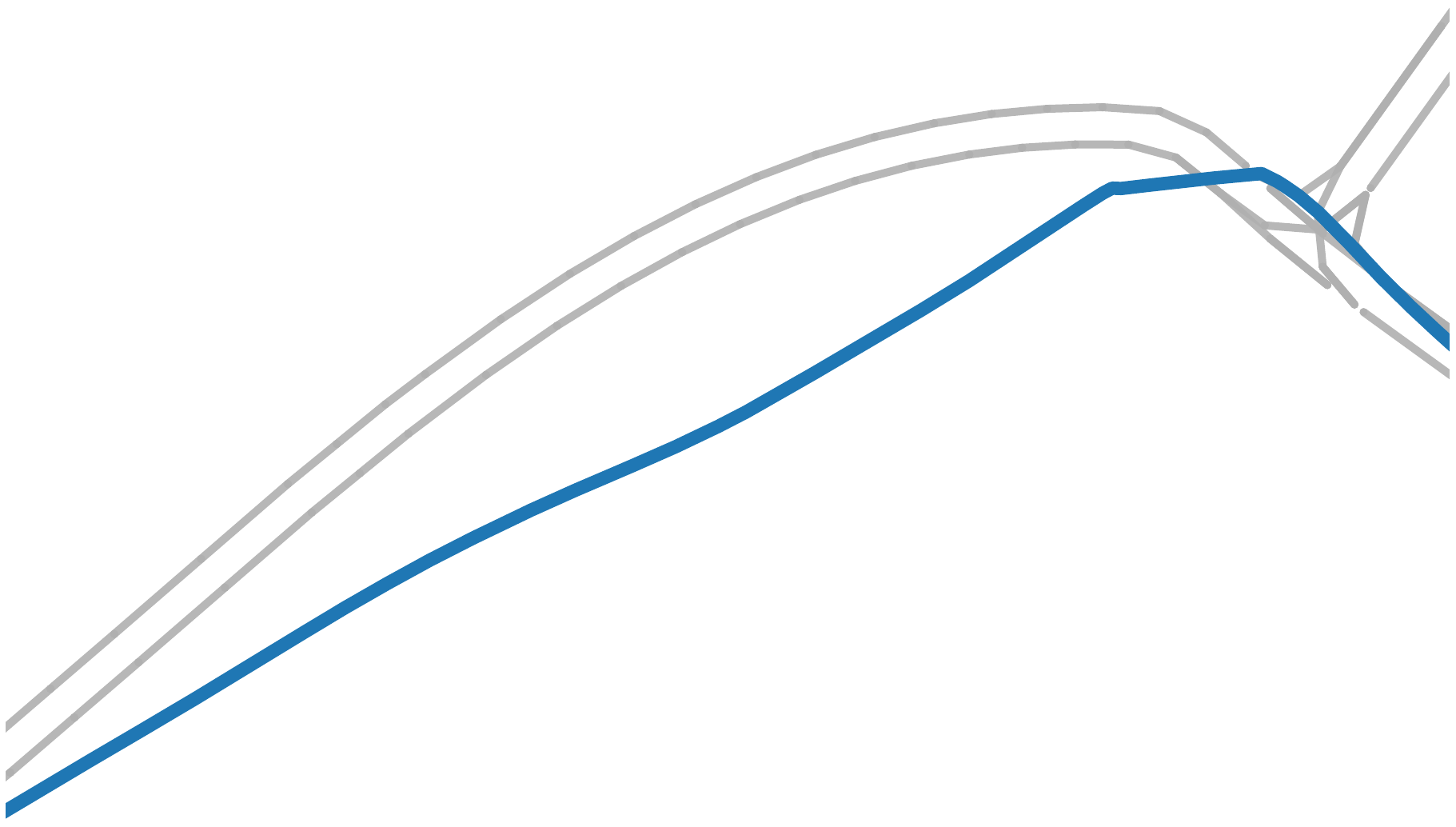}
    \caption{Comparison between KITTI ground truth (blue) and OSM-derived lane centerlines (gray) for sequence 00, illustrating a visible misalignment between the two input sources.}
    \label{fig:lanlet_osm_discrepancy}
    \end{figure}

    While conceptually simple, this process is challenging in practice due to several real-world issues: (i) lane maps may contain inaccuracies or missing parts, as shown in Fig.~\ref{fig:lanlet_osm_discrepancy}; (ii) roads often contain multiple parallel lanes and intersections that introduce ambiguous matches; (iii) odometry drift can become large, making data association difficult; and (iv) the map region can be very large, requiring efficient nearest-neighbor queries and filtering for real-time operation. For these reasons, the main technical components of our method specifically address efficient map representation, robust lane correspondence selection, and drift correction through enhanced ICP on a sliding window. These issues are addressed in the next sections.

    \subsection{Map Data Extraction and Representation}
    \label{subsec:map_data}

    We leverage OSM, an open-source geographic database containing road-level information on a global scale~\cite{OpenStreetMap}. Although OSM data may be noisier than proprietary HD maps, it provides wide coverage and is sufficient to reduce long-term drift by acting as a global geometric prior.

    Raw OSM data contain many objects unrelated to driving lanes. Therefore, we extract lane-level data through a lightweight preprocessing pipeline. First, road-related elements are queried and filtered using tools such as Overpass~\cite{overpassturbo}. The resulting OSM road geometry is then converted into the Lanelet2~\cite{poggenhans2018lanelet2} representation using CommonRoad~\cite{crdesingerMaierhofer2023} tools. Lanelet2 is a structured lane-level map format that represents roads as collections of points in a local Universal Transverse Mercator (UTM) coordinate frame, defined with respect to a chosen origin, allowing direct metric computations. It also provides derived centerlines from lane boundaries and encodes lane direction information. After conversion, we project the map into a local UTM coordinate system and precompute lane centerlines, which better approximate typical vehicle trajectories than lane boundaries. To control map resolution and runtime, we apply a minimum point-spacing filter of $3\,\mathrm{m}$ along each centerline, discarding consecutive points below this threshold. This removes redundant points in dense regions, such as curves, and reduces the number of KD-tree candidates used during matching. This spacing controls the trade-off between geometric resolution and computational cost: denser sampling can better preserve curved road geometry but increases the number of map points and nearest-neighbor queries, while overly sparse sampling may reduce runtime at the cost of less accurate lane matching.

    To support efficient matching at runtime, we represent the lane map as a set of centerline segments $\mathcal{S}=\{(\mathbf{a}_j,\mathbf{b}_j)\}$ and build a 2D KD-tree $\mathcal{K}$ over their sampled centerline points for fast nearest-neighbor retrieval. Each segment is associated with a normalized direction vector, which is used to reject candidates that are inconsistent with the vehicle motion direction.

    \subsection{Lane Correspondence Matching}
    \label{subsec:matching}

    The correction pipeline requires associating trajectory poses with corresponding lane locations. However, lane centerlines are typically sparsely sampled, and intersections or local map inaccuracies may introduce outliers. Therefore, a naive nearest-neighbor strategy based solely on the closest point is not sufficiently robust. Consequently, for each pose in the trajectory segment, we query the KD-tree to obtain the $N$ nearest candidate lane segments. The nearest-neighbor candidates are evaluated in increasing distance order, and the first candidate satisfying both constraints below is selected. This avoids an additional projected-distance search while preserving a simple and efficient association rule.
    
    \emph{(1) Normal projection constraint}. We project the trajectory point onto the local lane segment using \emph{normal shooting}. More precisely, we compute the intersection between the normal line passing through the trajectory point and the lane segment defined by two consecutive lane points; this provides a geometric constraint. Given a trajectory point $\mathbf{x}_i$, its unit tangent
    $\mathbf{d}_{\mathrm{traj}}$, and a candidate lane segment
    $(\mathbf{a}_j,\mathbf{b}_j)\in\mathcal{S}$, we define the trajectory
    normal as
    \[
    \mathbf{n}_i =
    \begin{bmatrix}
    -d_{\mathrm{traj},y} & d_{\mathrm{traj},x}
    \end{bmatrix}^{\top}.
    \]
    The intersection with the candidate lane segment is computed as
    \begin{equation}
    t =
    \frac{(\mathbf{x}_i-\mathbf{a}_j)\times\mathbf{n}_i}
         {(\mathbf{b}_j-\mathbf{a}_j)\times\mathbf{n}_i},
    \qquad
    \mathbf{q} =
    \mathbf{a}_j+t(\mathbf{b}_j-\mathbf{a}_j),
    \label{eq:normal_projection}
    \end{equation}
    where $\times$ denotes the two-dimensional cross product. The
    intersection is valid only if the denominator is nonzero and
    $0\leq t\leq1$.
    
    \begin{figure}[!b]
        \centering
        \includegraphics[width=0.65\linewidth, origin=c]{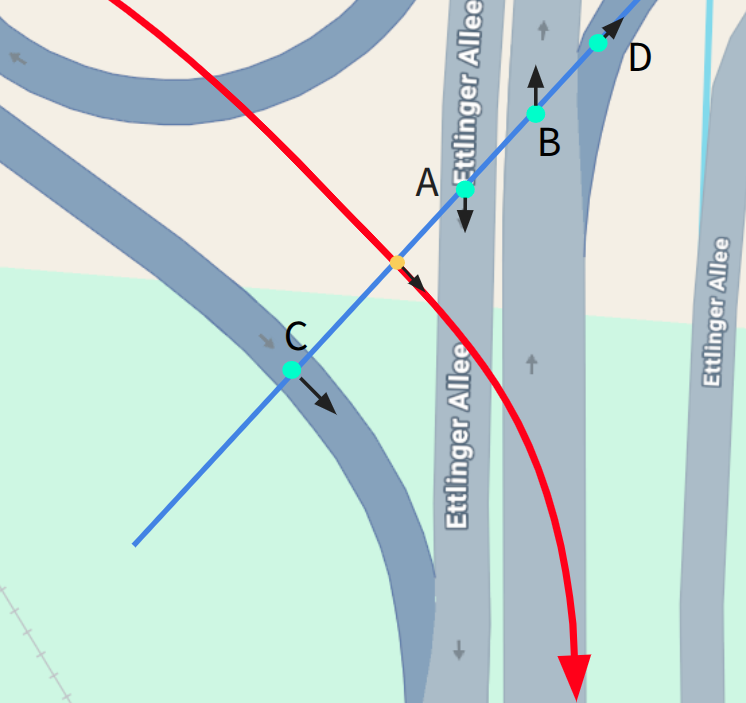}
        \caption{Illustration of lane–pose matching. \change{The red curve represents the trajectory, the yellow point is the pose under evaluation, the blue segment represents the normal projection direction,} and light-blue points correspond to valid candidate correspondences obtained from the intersection between the projection and neighboring lane segments. Finally, point C is selected as the correct matching point.}
        % \caption{Lane--pose matching. The red curve is the trajectory, the yellow point the current pose, cyan points nearby lane points, and the blue segment the normal projection. Light-blue points are valid candidates, with C selected as the final correspondence.}
        \label{fig:method_visualizer}
    \end{figure}

    If the projection lies within the lane segment bounds, it is considered a valid candidate correspondence. Fig.~\ref{fig:method_visualizer} illustrates the process: the evaluated pose (yellow) is projected along its normal direction onto nearby lane segments, producing multiple valid intersection candidates (light-blue), from which the geometrically consistent match is selected.

    \emph{(2) Directional Consistency Constraint}. We compute the trajectory tangent direction from consecutive poses and the lane direction from the candidate lane segment, as defined in \eqref{eq:directions}.  As illustrated in Fig.~\ref{fig:method_visualizer}, after applying the normal projection constraint (1), multiple valid candidates (A, B, C, D) may remain. To disambiguate them, we compare the trajectory direction with the direction of each lane segment: 
    
    \begin{align}
    \mathbf{d}_{\text{traj}} &=
    \frac{\mathbf{x}_{i+1} - \mathbf{x}_i}
    {\|\mathbf{x}_{i+1} - \mathbf{x}_i\|}, \quad
    \mathbf{d}_{\text{lane}} =
    \frac{\mathbf{b}_j-\mathbf{a}_j}
         {\|\mathbf{b}_j-\mathbf{a}_j\|}
    \label{eq:directions} \\[4pt]
    s &= \mathbf{d}_{\text{traj}}^\top \mathbf{d}_{\text{lane}},
    \label{eq:dotproduct}
    \end{align}
    where a correspondence is accepted only if $s\geq\tau$; in all
    experiments, $\tau=0.95$. This ensures that matches are selected on lane segments aligned with the vehicle's direction of travel. Since the candidate segments are evaluated in increasing nearest-neighbor distance, the first candidate satisfying both the projection and direction constraints is selected as the match. Following the example of Fig.~\ref{fig:method_visualizer}, the final candidate would be lane point C. If no candidate satisfies both constraints, no correspondence is added for that pose. Algorithm~\ref{alg:lane_matching} summarizes the approach.
    
    \begin{algorithm}[!b]
    \caption{Lane Point Matching via Normal Projection and Direction Filtering}
    \label{alg:lane_matching}
    \begin{algorithmic}[1]
    \REQUIRE Trajectory poses $\mathcal{P}=\{p_i\}_{i=1}^{M}$, lane segments $\mathcal{S}=\{(a_j,b_j)\}$, KD-tree $\mathcal{K}$ over lane segment points, number of neighbors $N$, direction threshold $\tau$
    \ENSURE Valid pose--lane correspondences $\mathcal{C}$
    
    \STATE $\mathcal{C} \leftarrow [\ ]$
    \FOR{$i = 1$ to $M$}
        \STATE $p \leftarrow p_i$
        \STATE $\mathbf{d}_{\text{traj}} \leftarrow \textsc{ComputeDirection}(p_i,p_{i+1})$
        \COMMENT{eq. \eqref{eq:directions}, Backward difference for $i=M$}
        \STATE $\mathcal{I} \leftarrow \mathcal{K}.\textsc{KNN}(p,N)$
        \COMMENT{$N$ candidate lane segments from $\mathcal{S}$ closest to $p$, sorted by distance}
    
        \FOR{each $j \in \mathcal{I}$}
            \STATE $(a_j,b_j) \leftarrow \mathcal{S}[j]$
            \STATE $\mathbf{d}_{\text{lane}} \leftarrow \textsc{ComputeDirection}(a_j,b_j)$
            \STATE $s \leftarrow \mathbf{d}_{\text{traj}}^\top \mathbf{d}_{\text{lane}}$
            \COMMENT{eq. \eqref{eq:directions} and \eqref{eq:dotproduct}}
    
            \IF{$s \geq \tau$}
                % \STATE $q \leftarrow \textsc{ProjectNormal}(p,a_j,b_j)$}                \COMMENT{eq. \eqref{eq:normal_projection}}
                \STATE $q \leftarrow
                    \textsc{ProjectNormal}(p,\mathbf{d}_{\text{traj}},a_j,b_j)$
                        \COMMENT{eq. \eqref{eq:normal_projection}}
                \IF{$q$ is valid}
                    \STATE Append $(p,q)$ to $\mathcal{C}$
                    \STATE \textbf{break}
                \ENDIF
            \ENDIF
        \ENDFOR
    \ENDFOR
    \RETURN $\mathcal{C}$
    \end{algorithmic}
    \end{algorithm}

    \subsection{Sliding-Window Trajectory Correction with Trimmed ICP}
    
    This stage constitutes the core of the proposed online correction pipeline, as summarized in Algorithm~\ref{alg:overall_pipeline}. The method processes odometer pose estimates sequentially and incrementally refines them using lane-based alignment within a sliding window framework.

    For each incoming pose $p_i$, the current accumulated correction transform $\Delta T_{\text{acc}} \in \text{SE}(2)$ is first applied to obtain a corrected pose $p_i^{\text{corr}}$. This pose is appended to the sliding window $\mathcal{W}$. The window stores the most recent corrected poses and is used to estimate local geometric alignment.

    Alignment is triggered only when the window contains at least $M_{\min}$ poses and spans a trajectory length of at least $L_{\min}$. If the window grows beyond the maximum size $M_{\max}$, the oldest pose is removed. This mechanism prevents overly long segments, which may introduce accumulated noise and degrade alignment quality, while maintaining bounded computational cost.

    Lane correspondences are computed using the \textsc{LaneMatch} routine,
    which associates poses in $\mathcal{W}$ with nearby lane segments using
    the projection and directional constraints described in
    Algorithm~\ref{alg:lane_matching}. The resulting set
    $\mathcal{C}$ contains valid pose--lane pairs. If
    $|\mathcal{C}|/|\mathcal{W}|\geq\eta$, trimmed ICP estimates the 2D
    rigid transform $\Delta T$ after discarding a fraction $\rho$ of the
    largest residuals and returns the mean Euclidean correspondence error $e$ over the retained
    matches; otherwise, the alignment is rejected.

    If the alignment error satisfies $e < \epsilon$, the correction is accepted. The accumulated global transform is updated as
    \[
    \Delta T_{\text{acc}} \leftarrow \Delta T \cdot \Delta T_{\text{acc}},
    \]
    and the sliding window poses are updated accordingly. The consecutive failure counter $E$ is reset to zero.

    {If the alignment error is greater than or equal to the threshold, the correction is rejected. In this case, the oldest pose is removed from the sliding window and the consecutive failure counter $E$ is incremented. This shortens the window and discards potentially unreliable early associations, allowing the next alignment attempt to focus on more recent motion.
    
    When the number of consecutive failures exceeds $E_{\max}$, a restart mechanism is triggered}. The window is reduced to the most recent $(M_{\min}-1)$ poses and the failure counter is reset. This enables recovery from persistent mismatches, temporary map inconsistencies, or severe drift.
    
    Overall, this sliding-window strategy allows for continuous correction while focusing on local motion consistency, maintaining robustness to noise, and ensuring bounded computational complexity.

    \begin{algorithm}[!b]
    \caption{Online OSM Lane-Based Drift Correction with Sliding Window and Trimmed ICP}
    \label{alg:overall_pipeline}
    \begin{algorithmic}[1]
    \REQUIRE Trajectory pose estimates $\mathcal{P}=\{p_i\}_{i=1}^{M}$
    \REQUIRE Lane segments $\mathcal{S}$, KD-tree $\mathcal{K}$
    \REQUIRE Minimum segment size $M_{\min}$, maximum segment size $M_{\max}$, trimming ratio $\rho$, ICP error threshold $\epsilon$, maximum consecutive failures $E_{\max}$, minimum correspondence ratio $\eta$, minimum trajectory length $L_{\min}$
    \ENSURE Corrected output poses $\mathcal{P}^{\text{corr}}$
    
    \STATE $\Delta T_{\text{acc}} \leftarrow I$
    \COMMENT{Accumulated correction in SE(2)}
    \STATE $\mathcal{W} \leftarrow [\ ]$
    \COMMENT{Sliding window of corrected poses}
    \STATE $\mathcal{P}^{\text{corr}} \leftarrow [\ ]$
    \COMMENT{Corrected output trajectory}
    \STATE $E \leftarrow 0$
    \COMMENT{Consecutive failure counter}
    
    \FOR{each pose $p_i$ in $\mathcal{P}$}
        \STATE $p_i^{\text{corr}} \leftarrow \Delta T_{\text{acc}} \cdot p_i$
        \STATE Append $p_i^{\text{corr}}$ to $\mathcal{W}$

        \IF{$|\mathcal{W}| > M_{\max}$}
            \STATE Remove oldest pose from $\mathcal{W}$
        \ENDIF
        \IF{$|\mathcal{W}| < M_{\min} \OR
        \textsc{TrajLength}(\mathcal{W}) < L_{\min}$}
            \STATE Append $p_i^{\text{corr}}$ to $\mathcal{P}^{\text{corr}}$
            \STATE \textbf{continue}
        \ENDIF
        \STATE $\mathcal{C} \leftarrow
        \textsc{LaneMatch}(\mathcal{W},\mathcal{S},\mathcal{K})$
        
        \IF{$|\mathcal{C}|/|\mathcal{W}| < \eta$}
            \STATE $(\Delta T,e) \leftarrow (I,\infty)$
            \COMMENT{Insufficient valid correspondences}
        \ELSE
            \STATE $(\Delta T,e) \leftarrow
            \textsc{TrimmedICP}(\mathcal{C},\rho)$
        \ENDIF

        \IF{$e < \epsilon$}
            \STATE $\Delta T_{\text{acc}} \leftarrow \Delta T \cdot \Delta T_{\text{acc}}$
            \STATE $\mathcal{W} \leftarrow \Delta T \cdot \mathcal{W}$
            \STATE $p_i^{\text{corr}} \leftarrow \Delta T \cdot p_i^{\text{corr}}$
            \STATE $E \leftarrow 0$
        \ELSE
            \STATE Remove oldest pose from $\mathcal{W}$
            \COMMENT{Discard potentially unreliable window start}
            \STATE $E \leftarrow E+1$
        
            \IF{$E > E_{\max}$}
                \STATE $\mathcal{W} \leftarrow$ keep last $(M_{\min}-1)$ poses
                \COMMENT{Restart matching from a shorter recent window}
                \STATE $E \leftarrow 0$
            \ENDIF
        \ENDIF
    
        \STATE Append $p_i^{\text{corr}}$ to $\mathcal{P}^{\text{corr}}$
    
    \ENDFOR
    
    \RETURN $\mathcal{P}^{\text{corr}}$
    \end{algorithmic}
    \end{algorithm}

\section{Experimental Setup}
 \label{sec:experiments}
    This section describes the experimental protocol used to evaluate the proposed OSM lane-based odometry correction method. We evaluate the method in two complementary ways. First, we apply it to multiple baseline odometry backends to demonstrate that the correction is independent of the underlying odometry source. Second, we compare against recent OSM-assisted drift-correction methods on KITTI Odometry, where published results are available for common baselines.

    \subsection{Baseline Odometry Estimation}
    \label{subsec:baseline}
        
    We evaluate the proposed correction on four odometry backends covering both LiDAR and visual odometry. KISS-ICP~\cite{kissicp} and ORB-SLAM3~\cite{orbslam3} are considered, respectively, state-of-the-art LiDAR and visual odometry/simultaneous localization and mapping (SLAM) baselines. They are included in the comparison because recent OSM-assisted LiDAR drift-correction methods~\cite{11367659,rl1a-kurda2024reducing} report results using KISS-ICP, while TOM-Odometry~\cite{9940585} reports results using ORB-SLAM3. We run all odometry backends on all evaluated datasets using their publicly available implementations and vanilla/default configurations, following the authors' instructions whenever possible. Therefore, results may differ from values reported in the original papers due to differences in evaluation protocol, metric definition, software version, sensor data preprocessing, and initialization. We also evaluate Basalt~\cite{basalt} as an additional state-of-the-art visual odometry backend, and LiODOM~\cite{garcia2022liodom} as an additional LiDAR odometry backend with more pronounced drift, providing a useful stress case for evaluating the robustness of the proposed correction.

    \subsection{Datasets}
    \label{subsec:datasets}
    We evaluate on two widely used autonomous driving benchmarks:

    \paragraph{KITTI Odometry}
    The KITTI Odometry~\cite{kittiodomGeiger2012CVPR} dataset provides synchronized stereo images, LiDAR point clouds, IMU data, and accurate ground-truth poses for multiple driving sequences. It includes suburban roads, highways, and urban scenarios, making it suitable for evaluating long-term drift accumulation. We exclude sequence~3 from evaluation due to missing Global Positioning System (GPS) data required to initialize and align the OSM map.

    \paragraph{KITTI-360}
    KITTI-360~\cite{kitti360} extends KITTI to longer trajectories and larger-scale urban environments, including challenging loop-free segments with pronounced drift. This dataset is especially relevant for evaluating map-assisted odometry correction. We exclude sequence~8, as it does not provide valid GPS data required to initialize and align the OSM map.

    \subsection{Comparison Methodology}
    \label{subsec:comparison_protocol}

    We compare against three OSM-assisted odometry correction/localization methods. The two most recent LiDAR--OSM methods~\cite{11367659,rl1a-kurda2024reducing} are compared using KISS-ICP~\cite{kissicp} as the common baseline. We also compare with TOM-Odometry~\cite{9940585}, since it is the closest prior work to ours in spirit: it uses OSM-derived road topology to correct odometry drift. Following the baseline reported by that work, TOM-Odometry is compared using ORB-SLAM3~\cite{orbslam3}. Since prior methods report results only for selected KITTI sequences, we evaluate on the sequences common to all methods and report both per-sequence two-dimensional absolute pose error (APE-2D) and the mean across them.

    \subsection{Evaluation Metric}
    \label{subsec:metric}
    
    We evaluate trajectory accuracy using evo-style no-alignment absolute errors~\cite{grupp2017evo}, since the goal of the proposed method is to reduce global drift directly in the map/ground-truth frame. Our primary metric is the mean absolute pose error in the horizontal plane, denoted APE-2D, because the correction operates in $\mathrm{SE}(2)$ and only refines $xy$ motion:
    
    \begin{equation}
    \mathrm{APE\text{-}2D}
    =
    \frac{1}{K}
    \sum_{i=1}^{K}
    \left\|
    \mathbf{t}_i^{xy}
    -
    \left(\mathbf{t}_i^{xy}\right)^{\mathrm{gt}}
    \right\|_2,
    \end{equation}
    where $K$ is the number of evaluated poses.
  
    Following the evo evaluation convention, we additionally report translational root mean square error (RMSE), lateral RMSE, and heading error under the same no-alignment setting. We do not use Relative Pose Error as the main metric because the sliding-window correction applies discrete rigid updates that may introduce local discontinuities, even when global trajectory consistency is improved.

    \subsection{Parameters and Implementation Details}
    \label{subsec:params}
    
    The correction algorithm is kept unchanged for all odometry backends. However, since each backend has different drift characteristics, we select its hyperparameters through a bounded parameter sweep. The swept parameters include the minimum and maximum sliding-window sizes, number of nearest-neighbor lane candidates, ICP acceptance threshold, and reset threshold.
    
    The minimum and maximum sliding-window sizes, $M_{\min}$ and
    $M_{\max}$, control when correction starts and how many recent poses
    are used for alignment. The number of nearest-neighbor lane candidates
    $N$ controls the size of the KD-tree search during correspondence
    matching. The ICP error threshold $\epsilon$ determines whether a
    computed correction is accepted or rejected, while $E_{\max}$ controls
    the number of consecutive failed alignments allowed before resetting
    the window. The trimming ratio is fixed at $\rho=0.1$. The correspondence ratio threshold is fixed at $\eta=0.5$, requiring at least 50\% of the poses to have a valid lane correspondence, and the minimum trajectory-length threshold is fixed at $L_{\min}=10\,\mathrm{m}$.
    
    For LiODOM, we use $M_{\min}=50$, $N=100$, $E_{\max}=50$, $\epsilon=2.0$\,m and $M_{\max}=3000$; for Basalt, we use $M_{\min}=150$, $N=10$, $E_{\max}=10$, $\epsilon=1.0$\,m and $M_{\max}=3000$; for KISS-ICP, we use $M_{\min}=150$, $M_{\max}=1000$, $N=100$, $E_{\max}=1000$, and $\epsilon=1.5$\,m; for ORB-SLAM3, we use $M_{\min}=150$, $M_{\max}=1000$, $N=50$, $E_{\max}=100$, and $\epsilon=1.5$\,m; for all backends, the directional consistency threshold is set to $\tau=0.95$.
    
    \subsection{Implementation}
    \label{subsec:implementation}

    The correction module is implemented in Python. Lane geometry is indexed with a KD-tree, and trimmed ICP~\cite{trimmedicp} uses standard NumPy/SciPy linear algebra routines; no GPU acceleration is employed. All experiments were run on a desktop machine equipped with an Intel Core i7-14650HX processor and 32\,GB of RAM, running Linux.

  \section{Results}
    \label{sec:results}
        
    In this section, we evaluate the proposed correction approach in three stages. First, we compare against prior OSM-assisted odometry-correction methods on the KITTI sequences commonly reported in the literature. Second, we evaluate the same correction framework across multiple odometry backends and datasets to show that the method is independent of the underlying odometry source. Finally, we report the computational cost of the correction module.

    \subsection{Comparison with OSM-Assisted Odometry Correctors}
    
    Table~\ref{tab:kiss_prior_comparison} compares our correction with the LiDAR--OSM drift-correction methods of Li et al.~\cite{11367659} and Kurda et al.~\cite{rl1a-kurda2024reducing}, using KISS-ICP as the common odometry input. Although no method dominates every sequence, the proposed method achieves the best mean over the full set of evaluated sequences and the lowest mean APE-2D on the subset \change{where Li et al. report results.}

    A sequence-dependent limitation is visible on KITTI sequence~01. Although the proposed method slightly improves KISS-ICP from $2.97\,\mathrm{m}$ to $2.81\,\mathrm{m}$ APE-2D, it does not outperform \change{Kurda et al.~\cite{rl1a-kurda2024reducing}} on this sequence. This is mainly due to the highway-like structure of the sequence, with long nearly straight segments and frequent lane changes. In such cases, lane-centerline matching provides weak geometric constraints, and parallel lanes or small OSM/georeferencing offsets can make the closest-lane association ambiguous.
    
    \begin{table}[!t]
    \caption{APE-2D comparison against KISS-ICP-based OSM correction methods for KITTI sequences. Lower is better.}
    \label{tab:kiss_prior_comparison}
    \centering
    \scriptsize
    \setlength{\tabcolsep}{2.5pt}
    \renewcommand{\arraystretch}{1.05}
    % \begin{tabular}{c r r r r}
    \begin{tabular}{c r r r @{\hspace{10pt}} r}
    \toprule
      Seq. & KISS-ICP~\cite{kissicp}
    & \change{Kurda et al.~\cite{rl1a-kurda2024reducing}}
    & \change{Li et al.~\cite{11367659}}
    & \change{Ours} \\
    \midrule
    00 & 3.54 & 1.46 & 2.97 & \textbf{1.31} \\
    01 & 2.97 & \textbf{1.57} & -- & 2.81 \\
    02 & 12.95 & 8.05 & -- & \textbf{2.23} \\
    04 & 0.47 & 0.43 & -- & \textbf{0.37} \\
    05 & 1.45 & 1.06 & 1.21 & \textbf{0.89} \\
    06 & 1.57 & \textbf{0.76} & 1.20 & 1.20 \\
    07 & \textbf{0.38} & 0.62 & 0.84 & 1.10 \\
    08 & 10.51 & 4.88 & \textbf{1.80} & 2.10 \\
    09 & 1.79 & 2.31 & \textbf{0.96} & 1.04 \\
    10 & 1.79 & \textbf{0.68} & -- & 0.85 \\
    \midrule
    Mean & 3.74 & 2.18 & -- & \textbf{1.39} \\
    Mean$^\dagger$ & -- & -- & 1.50 & \textbf{1.27} \\
    \bottomrule
    \end{tabular}
    \parbox{\columnwidth}{%
    \scriptsize
    $^\dagger$ Mean over the common sequences reported by Li et al.%
    }
    \end{table}

    Table~\ref{tab:orb_tom_comparison} compares against TOM-Odometry~\cite{9940585}, the closest road-geometry-based method reporting ORB-SLAM3 results. The comparison is restricted to the five KITTI sequences reported by TOM-Odometry. Our method improves ORB-SLAM3 on all five sequences and achieves a lower mean APE-2D than TOM-Odometry, while avoiding road-graph initialization, topological matching, and Kalman filter-based map fusion.
    
    \begin{table}[!t]
    \caption{APE-2D comparison with TOM-Odometry using ORB-SLAM3 on selected KITTI sequences. Lower is better.}
    \label{tab:orb_tom_comparison}
    \centering
    \footnotesize
    \renewcommand{\arraystretch}{1.05}
    % \begin{tabular}{c r r r}
    \begin{tabular}{c r r @{\hspace{10pt}} r}
    \toprule
    Seq. & ORB-SLAM3~\cite{orbslam3} & TOM-Odometry~\cite{9940585} & \change{Ours} \\
    \midrule
    00 & 5.28 & 1.30 & \textbf{1.29} \\
    02 & 5.97 & \textbf{1.70} & \textbf{1.70} \\
    05 & 1.40 & 2.30 & \textbf{1.23} \\
    08 & 7.79 & 2.30 & \textbf{2.29} \\
    09 & 3.14 & 3.10 & \textbf{2.42} \\
    \midrule
    Mean & 4.72 & 2.14 & \textbf{1.79} \\
    \bottomrule
    \end{tabular}
    \end{table}
    
    \subsection{Results Across Odometry Backends}

    Table~\ref{tab:backend_summary} summarizes the correction performance across KITTI and KITTI-360. The proposed method reduces the mean APE-2D for all evaluated odometry backends on both datasets, showing that the correction is not tied to a specific odometry source. The largest gains occur on KITTI-360, where longer trajectories produce stronger accumulated drift. For example, LiODOM decreases from $334.33$\,m to $19.24$\,m, while KISS-ICP decreases from $82.73$\,m to $12.17$\,m. The same trend is also observed for visual odometry, with Basalt and ORB-SLAM3 both improving on KITTI and KITTI-360.
    
    \begin{table}[!t]
    % \caption{Backend summary. Lower is better.}
    \caption{Backend summary. Mean, RMSE, and lateral errors are reported
in meters, while heading errors are reported in degrees. Lower is better.}
    \label{tab:backend_summary}
    \centering
    \scriptsize
    \renewcommand{\arraystretch}{1.05}
    \begin{tabular}{l l l r r r r}
    \toprule
    Data & Backend & Method & Mean & RMSE & Lat. & Head. \\
    \midrule
    KITTI & LiODOM & Raw  & 104.61 & 121.25 & 99.64 & \textbf{6.9} \\
          &        & Ours & \textbf{59.84} & \textbf{68.06} & \textbf{54.94} & 7.2 \\
          & Basalt & Raw  & 19.46 & 23.64 & 12.49 & \textbf{4.7} \\
          &        & Ours & \textbf{16.38} & \textbf{20.33} & \textbf{9.67} & 5.6 \\
          & KISS-ICP & Raw  & 3.74 & 9.40 & 6.06 & 2.1 \\
          &          & Ours & \textbf{1.39} & \textbf{1.99} & \textbf{1.51} & \textbf{2.0} \\
          & ORB-SLAM3 & Raw  & 5.34 & 8.39 & 5.34 & \textbf{1.8} \\
          &           & Ours & \textbf{3.32} & \textbf{5.77} & \textbf{3.12} & 2.4 \\
    \midrule
    KITTI-360 & LiODOM & Raw  & 334.33 & 432.45 & 327.42 & 45.9 \\
              &        & Ours & \textbf{19.24} & \textbf{29.66} & \textbf{20.91} & \textbf{8.8} \\
              & Basalt & Raw  & 30.87 & 41.57 & 28.25 & \textbf{5.7} \\
              &        & Ours & \textbf{17.60} & \textbf{25.76} & \textbf{16.01} & 6.5 \\
              & KISS-ICP & Raw  & 82.73 & 164.95 & 111.77 & 14.4 \\
              &          & Ours & \textbf{12.17} & \textbf{14.41} & \textbf{11.06} & \textbf{4.3} \\
              & ORB-SLAM3 & Raw  & 22.90 & 36.89 & 29.16 & 8.9 \\
              &           & Ours & \textbf{12.21} & \textbf{25.15} & \textbf{23.99} & \textbf{8.8} \\
    \bottomrule
    \end{tabular}
    \end{table}

\begin{figure}[!t]
    \centering
    \includegraphics[width=\columnwidth]{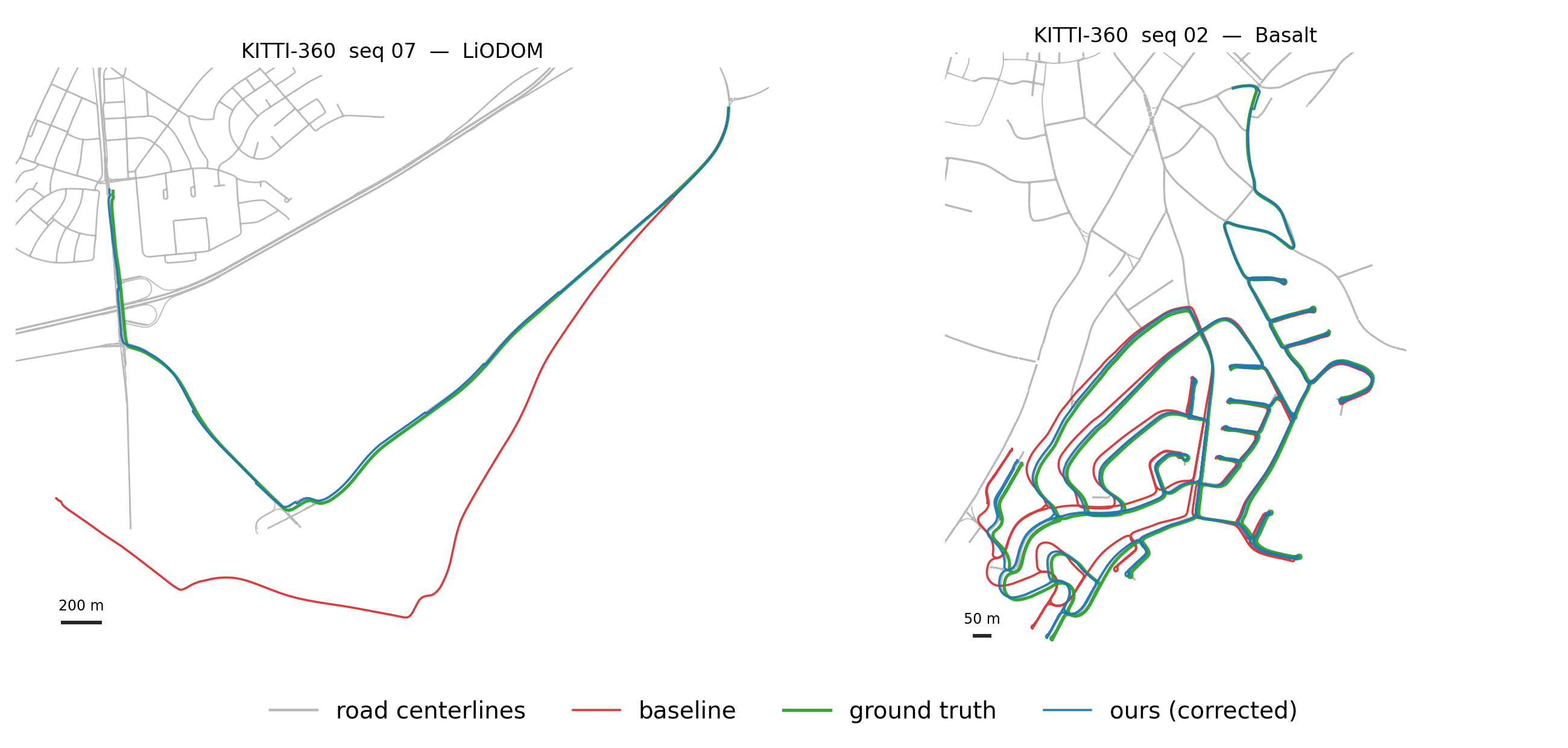}
    \caption{Representative KITTI-360 corrections using OSM lane centerlines. Left: LiODOM seq.~07. Right: Basalt seq.~02.}
    \label{fig:traj_overlay}
    \end{figure}

     Figure~\ref{fig:traj_overlay} further illustrates these backend corrections on challenging KITTI-360 sequences. In the LiODOM case, the raw trajectory accumulates severe drift, but the corrected trajectory is repeatedly pulled back toward the road layout. In the Basalt case, the odometry remains locally smooth but becomes globally displaced, and the proposed correction uses lane geometry as a lightweight global reference to recover consistency. These examples show that, even when odometry drifts globally, the local trajectory shape often preserves enough road-geometry information for alignment with OSM lane centerlines.
    
    \subsection{Computational Cost}
    \label{subsec:computational_cost}
    
    \change{Across all dataset--backend configurations, the correction requires $19.5$\,ms per pose on average ($51$\,Hz), with $19.1$\,ms for lane matching and $0.33$\,ms for trimmed ICP; total runtime ranges from $14.2$--$27.7$\,ms ($36$--$70$\,Hz). The KD-tree is built once per dataset ($39$\,ms for KITTI and $30$\,ms for KITTI-360), making the method compatible with typical $10$\,Hz LiDAR and $10$--$30$\,Hz camera odometry.}

\section{Conclusions}

    We presented an OSM lane-centerline correction method for reducing odometry drift without dense maps or odometer-specific assumptions. Experiments on KITTI and KITTI-360 show consistent improvements across LiDAR and visual odometry backends, including LiODOM, Basalt, KISS-ICP, and ORB-SLAM3. The method also remains competitive with recent OSM-assisted correctors while using a simpler and more general alignment strategy.
    
    The main limitations arise when lane association becomes ambiguous. Frequent lane changes or aggressive driving can reduce matching accuracy, even though the error counter and window reset mechanism help reject inconsistent corrections. Similarly, overlapping road structures such as bridges or tunnels are challenging when different road levels have similar 2D geometry and direction, since the current method does not explicitly model elevation or road-layer constraints. Future work will consider lane-change detection and vertical-level constraints to improve robustness in these cases. In addition, as discussed in Sec.~\ref{subsec:map_data}, OSM is not an HD map; missing lane counts, inaccurate lane widths, or shifted geometry can directly affect correspondence quality. Higher-fidelity lane maps or complementary perception cues, such as camera-based lane detection, could further reduce sensitivity to these map errors. Finally, because corrections are applied as discrete $\mathrm{SE}(2)$ updates, the resulting trajectory may contain local discontinuities; applications that require smooth motion, such as control or local planning, may therefore need an additional filtering or smoothing stage.

\section*{Acknowledgments}
    \change{This work has been partially supported by grant PID2022-139248NB-I00 (funded by MICIU/AEI/10.13039/501100011033 and by ERDF/EU). The work has been partially supported by the funding of the EMJMD master's program in Engineering of Data-Intensive Intelligent Software Systems (EDISS), issued by the European Union's Education, Audiovisual and Culture Executive Agency (grant number 619819). The authors acknowledge the support of Distance Technologies Oy.}
    
\bibliographystyle{IEEEtran}
\bibliography{ms}

\end{document}